\documentclass[letterpaper, 10 pt, conference]{ieeeconf}  % Comment this line out if you need a4paper
\IEEEoverridecommandlockouts
\usepackage{cite}
\usepackage{amsmath,amssymb,amsfonts}
\usepackage{graphicx}
\usepackage{textcomp}
\usepackage{xcolor}
\usepackage{cleveref}
\usepackage{bm}
\usepackage{bbm}
\usepackage{dsfont}
\usepackage{amsthm}
\usepackage{algorithm}
\usepackage{algpseudocode}
\usepackage{url}
\usepackage{multirow}
\usepackage{graphicx}    % for \includegraphics
\usepackage{subcaption}  % for subfigures
\usepackage{xcolor}    % for coloring
\usepackage{siunitx}   % for aligned numbers
\usepackage{booktabs}
\usepackage{adjustbox} % in your preamble
\usepackage{array} 
\usepackage[skip=5pt]{caption} % "skip" controls BOTH the space above and below the caption text.

\def\BibTeX{{\rm B\kern-.05em{\sc i\kern-.025em b}\kern-.08em
    T\kern-.1667em\lower.7ex\hbox{E}\kern-.125emX}}

\theoremstyle{definition}

\newtheorem{problem}{Problem}

\algnewcommand\algorithmicinput{\textbf{Input:}}
\algnewcommand\Input{\item[\algorithmicinput]}
\algnewcommand\algorithmicoutput{\textbf{Output:}}
\algnewcommand\Output{\item[\algorithmicoutput]}
\algnewcommand\algorithmicglobalvars{\textbf{Global Variables:}}
\algnewcommand\GlobalVars{\item[\algorithmicglobalvars]}

\DeclareMathOperator*{\argmax}{arg\,max}
\def \transpose{^\mathsf{T}}

\newcommand{\squeezeup}{\vspace{-1.5\baselineskip}}
\def\authorInfo{
    Leonard Jung$^1$, 
    Alan Papalia$^{1,2}$, 
    Kevin Doherty$^{3}$, 
    Michael Everett$^{1}$
    \thanks{%
        This research was funded by the DEVCOM Army Research Laboratory (ARL) in part by W911NF-24-2-006 and SARA CRA W911NF-24-2-0017
    }%
    \thanks{%
        $^{1}$Northeastern University, USA.
        $^{2}$University of Michigan.
        $^{3}$Boston Dynamics.
    }%
}
\author{%
    \authorInfo
}

\begin{document}

\title{Practical and Performant Enhancements for Maximization of Algebraic Connectivity}

\maketitle

\begin{abstract}
Long-term state estimation over graphs remains challenging as current graph estimation methods scale poorly on large, long-term graphs. To address this, our work advances a current state-of-the-art graph sparsification algorithm, maximizing algebraic connectivity (MAC). MAC is a sparsification method that preserves estimation performance by maximizing the algebraic connectivity, a spectral graph property that is directly connected to the estimation error. Unfortunately, MAC remains computationally prohibitive for online use and requires users to manually pre-specify a connectivity-preserving edge set. Our contributions close these gaps along three complementary fronts: we develop a specialized solver for algebraic connectivity that yields an average 2x runtime speedup; we investigate advanced step size strategies for MAC’s optimization procedure to enhance both convergence speed and solution quality; and we propose automatic schemes that guarantee graph connectivity without requiring manual specification of edges. Together, these contributions make MAC more scalable, reliable, and suitable for real-time estimation applications.

\textbf{Code: } To be released after paper acceptance
%\url{https://github.com/neu-autonomy/mac-pp}
\end{abstract}

% \begin{IEEEkeywords}
% SLAM, Graph Sparsification, 
% \end{IEEEkeywords}

\section{Introduction}
The scalability of state estimation and perception remains a critical challenge for long-term autonomous robotic systems. Many of these problems, such as simultaneous localization and mapping (SLAM), pose graph optimization (PGO), and structure from motion (SFM) may be formulated as an optimization problem over graphs. 
% Factor graphs provide an elegant and widely adopted framework for formulating these problems as maximum a posteriori (MAP) inference, represented as a nonlinear least-squares optimization.
However, as a system operates over an extended period, the graph may grow unchecked, accumulating a vast number of edges from odometry, loop closures, and other sensor measurements. This unbounded growth renders the underlying optimization problem computationally intractable, jeopardizing the real-time performance essential for autonomous operation.

One paradigm in which this issue is tackled is through \textit{graph sparsification}: the removal of edges while retaining the essential information within the graph. Instead of solving a large, complicated estimation problem over the entire graph, the goal is to approximate it with a sparser one for which the problem can be solved more efficiently. This naturally raises the question of what figure of merit defines a ‘good’ sparsified graph. One particularly important metric is the algebraic connectivity, which is closely tied to the expected performance of estimators \cite{9337203,rosen2019se}.

However, this edge selection problem is unfortunately NP-hard \cite{10.1016/j.orl.2008.09.001}. Recent work \cite{9981584} posed a convex relaxation that is more amenable to solving, but it remains computationally demanding. 
We build upon this framework, and improve both its computational efficiency and resulting graph quality.
First, we leverage the specific sparsity pattern inherent to the Laplacian graph for PGO graphs and develop a numerical conditioning strategy around this pattern. Secondly, we investigate  methods to specify/return a connected subgraph, the selection of which is critical to solution quality, yet has previously been unexplored.

\begin{figure}[t!]
    \centering
    \includegraphics[width=\linewidth]{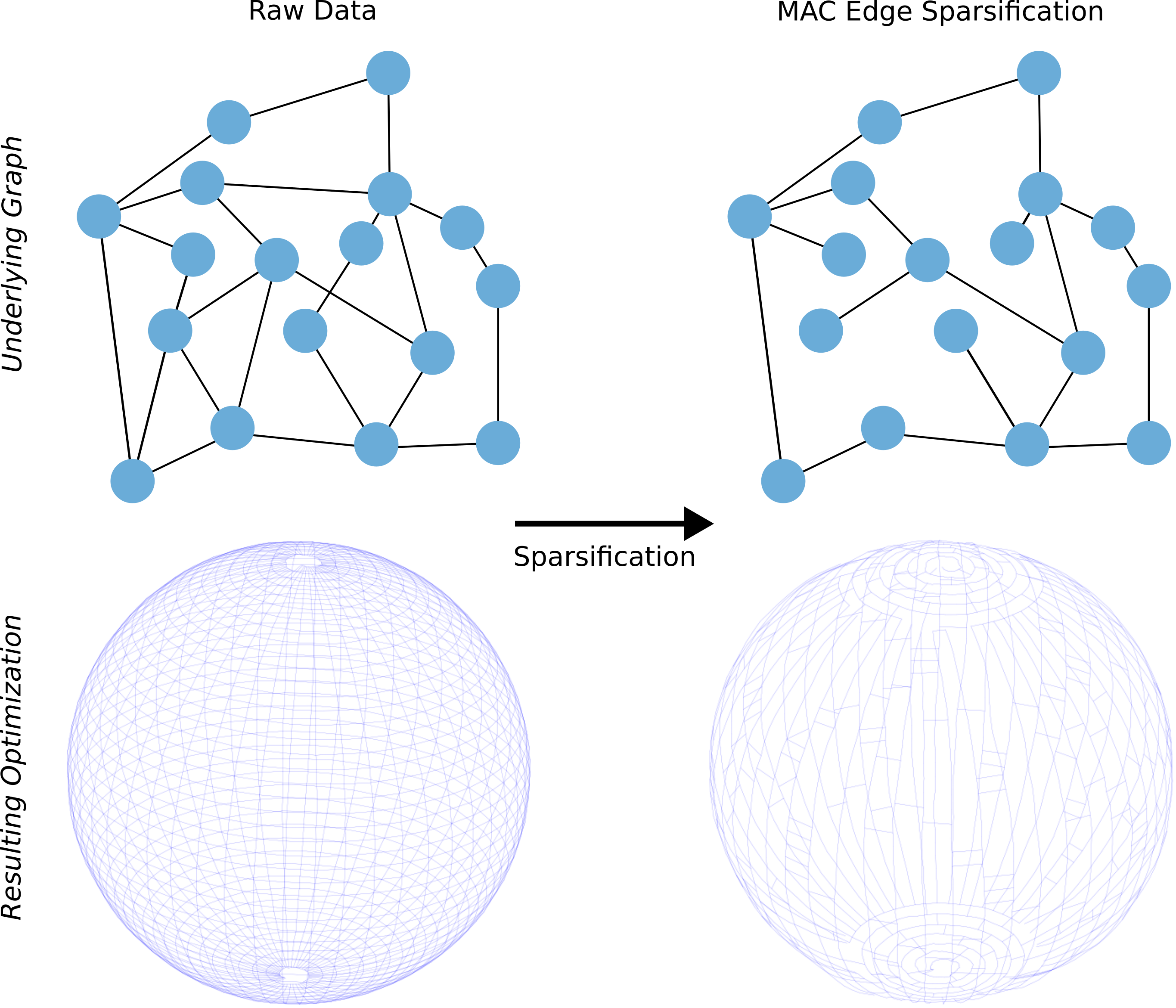}
\caption{We sparsify a graph by pruning edges in a way that maintains the graph’s structural properties, as quantified by its algebraic connectivity. (Top) Example of a ``mock" graph undergoing sparsification. (Bottom) Solution of the estimation problem on the sparsified graph retains the overall shape of the original while using significantly fewer edges.}
\label{fig:title}
% \vspace{-0.2in}
\squeezeup
\end{figure}

Our primary contributions are as follows: 
\begin{enumerate}
    \item We present an efficient strategy for computing the Fiedler value via Krylov-subspace methods and sparse matrix factorization, yielding a two-fold acceleration in solution times compared to a general-purpose approach.
    \item We investigate the impact of line-search strategies and other variants of the Frank-Wolfe algorithm on solving the maximizing algebraic connectivity problem, and empirically determine that the simple decaying step-size for Frank-Wolfe has the best convergence rate
    \item We present a method to remove previous (restrictive) assumptions that a backbone is given and present an improved heuristic for determining this (now optional) backbone

\end{enumerate}

\section{Related Works}
The pursuit of computational efficiency in robotics graph optimization problems has motivated a wide range of sparsification techniques. This section reviews related work, beginning with a broad overview of sparsification methods employed in perception and estimation. We then narrow our focus to a more formal class of graph-theoretic sparsification algorithms, within which our work on maximizing algebraic connectivity is situated. Finally, we discuss the key computational considerations for calculating the Fiedler value, a bottleneck in any iterative spectral sparsification routine. 

\subsection{Sparsification Techniques in Perception and Robotics}
Several methods address memory and computation costs for long-term autonomy using heuristics to sparsify nodes and edges such as \cite{6696478,kurz2021geometry, 6906954}. \cite{11127770,7989448} utilize submodular optimization to select which keyframes or features are most important to track, inherently building a smaller, more computationally efficient graph. Relatedly, point cloud sparsification techniques such as \cite{11127666} also reduce memory and computational constraints by subsampling sensor data prior to graph construction. We aim to directly solve a relaxed convex optimization problem on the underlying graph's spectral properties for sparsification, while maintaining suboptimality guarantees.

Tangentially, \cite{9423294} use spectral properties to specifically solve the PGO problem by leveraging sparsity; our method can be seen as a preprocessing step for any optimization problem over graphs.
\subsection{Fiedler Value Graph Sparsification}
Our work is most directly related to spectral graph sparsification methods that aim to preserve the algebraic connectivity (the Fiedler value) of the graph Laplacian. This line of research is rooted in the foundational work of \cite{Fiedler1973}, which established the importance of the Fiedler value as a measure of graph connectivity.

The MAC (maximizing algebraic connectivity) algorithm \cite{doherty2024mac} represents the state-of-the-art in Fiedler value-based graph sparsification. Our work identifies and addresses several of its key computational limitations to enhance its practicality.

The connection between algebraic connectivity and system performance has been extensively studied in theory and demonstrated empirically in multiple previous works.
 \cite{10333265} also uses the Fiedler value for sparsification and additionally computes the number of edges to remove. \cite{boyd2006convex} formulates the problem of adding edges to maximize the Fiedler value as a convex semidefinite program (SDP). Methods like OASIS \cite{10611644} also solve a similar smallest eigenvalue sparsification problem for sensor placement.
 
Like \cite{doherty2024mac}, it is important to distinguish our goal from that of spectral sparsification \cite{spielman2011spectral}, which aims to preserve the entire spectrum of the original graph Laplacian. Our objective is narrower: we seek only to maximize the Fiedler value. This is a less restrictive condition that allows for more aggressive sparsification and is directly related to the performance metrics of estimation problems, such as the Cramer-Rao lower bound \cite{9337203}.
\subsection{Fiedler Value Calculation}

A practical bottleneck in the MAC algorithm is the frequent computation of the Fiedler value and its corresponding eigenvector. Shift-invert Lanczos methods and other Krylov subspace techniques \cite{doi:10.1137/S0895479800371529} are both well-established methods for computing eigenvalues. The Tracemin-Fiedler \cite{10.5555/1964238.1964283} algorithm presented a method more tailored to finding the Fiedler value. As iterative solvers, the performance of these routines is heavily dependent on the conditioning of the matrix. Advanced vertex ordering (e.g., \cite{10.1145/1024074.1024081}) and preconditioning techniques \cite{9940527} can improve this. Our work uses several of these techniques along with properties of a graph Laplacian to drastically decrease computation time.

\section{Background}
\begin{figure*}[t!]
    \centering
    \includegraphics[width=0.9\linewidth]{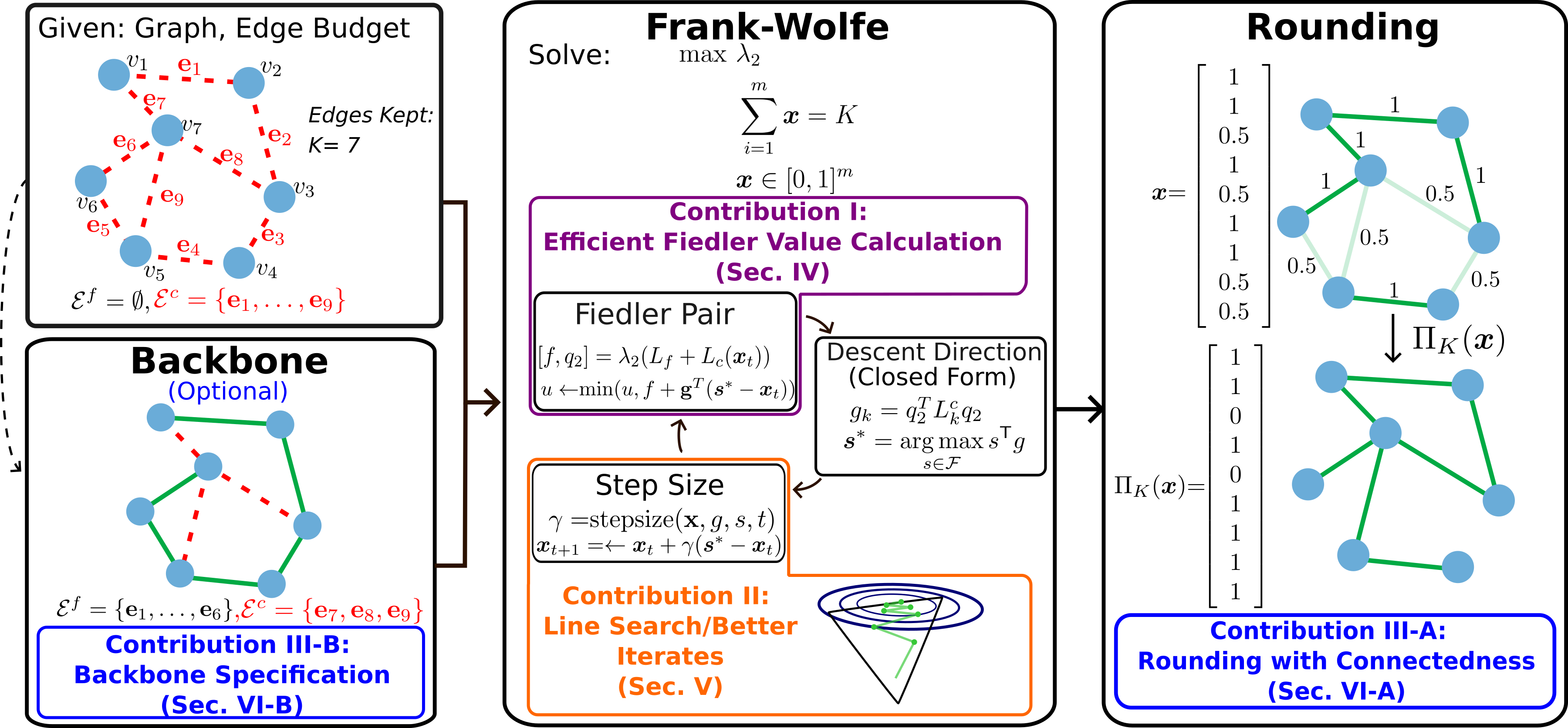}
\caption{\textbf{Overview of MAC algorithm with highlighted contributions:} MAC \cite{doherty2024mac} is given a graph and prespecified ``backbone" (fixed edge set), then applies a Frank–Wolfe procedure with a decaying step size, and finally rounds the solution to a binary one. To reduce total runtime and broaden MAC’s applicability to problems without a pre-specified backbone, we investigate three main questions: \textbf{1.} Can we increase MAC's runtime through Faster Fiedler value calculations (\Cref{sec:FiedlerValue})?; \textbf{2.} Can we increase MAC's convergence rate through intelligent step size and linesearch routines (\Cref{sec:lineSearch})?; \textbf{3.} Can we extend MAC to problems without a given fixed edge set through rounding functions $\Pi_K$(\Cref{sec:rounding}) and automatic backbone edge specification (\Cref{sec:effR})?}
\squeezeup
\label{fig:flow}
\end{figure*}

Many robotic perception problems are traditionally posed as estimation over graphs $\mathcal{G} \triangleq(\mathcal{V},\mathcal{E},w)$, where the variables of interest are the nodes $\mathcal{V}$, measurements are the edges $\mathcal{E}$, and precision are the edges $w$. 

Graphs found in robotics applications can easily exceed $10^5$ edges. Since estimation scales poorly in both memory and computational complexity with the number of edges, sparsification is crucial for enabling reasonable performance.
\subsection{Algebraic Connectivity}
The algebraic connectivity, also known as the Fiedler value, of a graph is the second smallest eigenvalue $\lambda_2$ of the Laplacian $L\in\mathbb{R}^{n\times n}$ of graph $\mathcal{G}$ where:
\begin{align*}
    L_{ij}=\begin{cases}
        \sum_{e\in\delta(i)}w_{e}, & i=j\\
        -w_{ij}. & \{i,j\}\in\mathcal{E}\\
        0 & \{i,j\}\notin\mathcal{E}
    \end{cases}    
\end{align*}
This graph Laplacian has several key properties of note:
\begin{itemize}
    \item It is always positive semi-definite (PSD).
    \item It has at least one zero eigenvalue, which corresponds to the ``all ones" eigenvector $\mathbf{1}$. More generally, the multiplicity of the zero eigenvalue equals the number of connected components in the graph
    \item The Laplacian can be written as the sum of the Laplacian of the subgraphs induced by each of its edges.
\end{itemize}
We refer to the eigenvector associated with the Fiedler Value as the Fiedler vector $q_2$ and the two together as the Fiedler Pair.

The algebraic connectivity is inversely related to the worst-case error for solutions of many estimation problems \cite{rosen2019se,doherty2024mac}; that is, if we aim to determine which states and measurements should be pruned while minimally changing the quality of our solution, the resulting graph should maximize the algebraic connectivity of the graph with nodes corresponding to robot poses $x_i$ and edge weights equal to $w_{ij}$.

\subsection{Maximization of Algebraic Connectivity}
Given an edge budget $K$, our objective is to find the set of $K$ edges that maximizes the algebraic connectivity (Fiedler value) of the sparsified graph. We formalize this as the following problem, where $L(\bm{x})=\sum_{k=1}^m \bm{x}_kL_k$:
\begin{problem} Fiedler Value Maximization
\begin{equation} \label{eq:orioginal_mac}
    \begin{gathered}
        \max_{\bm{x}\in\{0,1\}^{m}}\lambda_{2}\left(L(\bm{x})\right) \\\sum_{k=1}^{m}\bm{x}_{k}=K
     \end{gathered}
\end{equation}
\label{prob:originalKchoice}
\end{problem}
However, due to the integrality constraint $\bm{x}_k\in\{0,1\}$ this problem is NP-Hard. \cite{doherty2024mac} solves this via a convex relaxation:
\begin{problem} Convex Relaxation of Fiedler Value Maximization
    \begin{equation}
        \begin{gathered}
            \bm{x^*}=\argmax_{\bm{x}\in[0,1]^{m}} \lambda_{2}\left(L(\bm{x})\right)\\ 
            \sum_{k=1}^{m}\bm{x}=K \label{eq:og_eq}
         \end{gathered}
    \end{equation}
    \label{prob:relaxedKchoice}
\end{problem}

As the Fiedler value of a Laplacian is a concave function \cite{4177113} of $\bm{x}$, Problem \ref{prob:relaxedKchoice} is concave and can be efficiently solved to global optimality. However, due to the relaxation of the integrality constraint, the solution of Problem \ref{prob:relaxedKchoice} may not be a feasible point for Problem \ref{prob:originalKchoice}.

If the solution to Problem \ref{prob:relaxedKchoice} results in a non-integral solution, a rounding function is used to return a valid solution:
\begin{equation}\label{eq:rounding}
    \bm{x}_{\{0,1\}^m}^{*} = \Pi_K(\bm{x^*})
\end{equation}
where $\Pi_K:[0,1]^m\rightarrow\{0,1\}^m$ is any function that rounds $\bm{x}$ to an integral solution, and $\sum_k\bm{x}_k=K$. This rounding function can drastically impact the final graph: for instance, poor rounding may result in a disconnected graph. Therefore, the choice of rounding strategy plays a critical role in the final solution quality.

\subsection{Frank-Wolfe for Maximizing Algebraic Connectivity}
\cite{doherty2024mac} demonstrates that the Frank-Wolfe algorithm is particularly well-suited to solve Problem \ref{prob:relaxedKchoice}. The Frank-Wolfe algorithm consists of two steps. First, a descent direction is found from minimizing the linearized objective function
\begin{equation}
    s=\arg\min_{s\in\mathcal{F}} \langle\nabla f(\bm{x}^{(t)}),\bm{s}\rangle
\end{equation}
Then, the next iterate is found by taking a step in the descent direction $\bm{x}^{(t+1)}=\bm{x}^{(t)}+\gamma(\bm{s}-\bm{x}^{(t)})$. Although the Frank-Wolfe algorithm is designed for smooth convex functions, it is easily adapted to non-smooth, concave functions such as the Fiedler Value by negating and using sub-gradients. 

For Problem \ref{prob:relaxedKchoice}, each Frank-Wolfe iteration requires only a single evaluation of the Fiedler value and its corresponding eigenvector. A super-gradient (the concave analog to the sub-gradient) $g\in\mathbb{R}^m$ of $\lambda_2(\bm{x})$ can then be found:
\begin{equation} \label{eq:supergradient}
    g = (I_{ m }\otimes q_{2})\transpose \begin{pmatrix} L _ { 1 } ^ { c } \\ \vdots \\ L _ { m } ^ { c } \end{pmatrix} q _ { 2 }
\end{equation}
where $q_2$ is a Fiedler vector and $I_m$ is the identity matrix.
% Of note as, $\lambda_2$ is a concave function, $g$ is technically a \textit{super-gradient}

The descent direction $\bm{s}=[\bm{s}_1,\dots,\bm{s}_m]\transpose$problem is then solved in closed form
\begin{equation} \label{eq:directionfinding}
    \bm{s}_k=\begin{cases}1, & k\in\mathcal{S}^{*},\\ 0, & \text{otherwise},\end{cases}
\end{equation}
where $S^*$ corresponds to the indices of $g$ with the $K$ largest entries.
% Finally, the dual of (\ref{eq:og_eq}) allows for quantifying the sub-optimally post-hoc of optimization.
\subsection{MAC Algorithm}
% \TODO{ADD a picture/diagram/algorithm of the MAC algorithm}
\begin{algorithm}[t!]
\caption{MAC Algorithm \cite{doherty2024mac}}
\label{alg:mac}
\begin{algorithmic}[1]
\Input $\mathcal{E}^f\gets$ Fixed Edges, $\mathcal{E}^c \gets$ Candidate Edges, $T\gets$ Total Iterations, $\epsilon_u\gets$ Duality Gap, $\bm{x}^{(0)}\gets$ Initial Guess, $K\gets$ Number of Candidate Edges to Keep
% \State \textbf{Initialize:} $L_f\gets$ Fixed Laplacian, $\bm{x}_1\in \{0,1\}^m\gets $ Initial Guess
\Output Edges Kept
\Function{MAC}{$\mathcal{E}^f, \mathcal{E}^c,T,\, \bm{x}^{(0)}$}
    % \State Define \(g: v \mapsto(\ref{eq:supergradient})\)
    \For{$t = 0, \ldots, T-1$} \Comment{Frank-Wolfe}
        \State $f \gets \lambda_2(L_f+\sum_{k=1}^m \bm{x}^{(t)}_kL_k^c)$
        % \State $\bm{g} \gets (I_{m} \otimes q_{2}(\bm{x}_t))\transpose \begin{pmatrix} L_{1}^{c} \\ \vdots \\ L_{m}^{c} \end{pmatrix} q_{2}(\bm{x}_t)$
        % \State $\bm{s}^* \gets \arg\max_{\bm{s}\in\mathcal{F}} s^\transpose \bm{g}$
        \State $\bm{g},\bm{s} \gets$ from (\ref{eq:supergradient}) and (\ref{eq:directionfinding}), respectively
        \State $u \gets \min(u, f + \bm{g}\transpose (\bm{s} - \bm{x}^{(t)})$ \Comment{Dual}
        \If{$|\frac{u-f}{f}| < \epsilon_u$}
            \State \text{break}
        \EndIf
        \State $\gamma \gets \frac{2}{2+t}$ \Comment{Naive Step Size}
        \State $\bm{x}^{(t+1)} \gets \bm{x}^{(t)}+\gamma(\bm{s}-\bm{x}^{(t)})$
    \EndFor
    \State \Return $\Pi_K(\bm{x}^{(T)})$ \Comment{Madow Sampling Rounding}
\EndFunction
\end{algorithmic}
\end{algorithm}
The MAC algorithm \cite{doherty2024mac} is the leading method for scalable sparsification and serves as our primary foundation.

It first assumes sets of fixed edges, forming a connected subgraph $\mathcal{E}^f$, and  candidate edges $\mathcal{E}^c$ are given, where $\mathcal {E}=\mathcal{E}^f \cup \mathcal{E}^c, \mathcal{E}^f \cap \mathcal{E}^c=\emptyset$. Next, using the Frank-Wolfe Algorithm, it solves Problem \ref{prob:relaxedKchoice}, with the Laplacian rewritten in terms of the edge sets and selection vector $\bm{x}$
\begin{equation} \label{eq:Lmap}
    L(\bm{x})= \lambda_2(L_f+\sum_{k=1}^m \bm{x}_kL_k^c)
\end{equation}
where $L_f$ and $L_k^C$ are the Laplacians corresponding to edges in $\mathcal{E}^f$ and edge $e_k\in\mathcal{E}^c$, respectively. Finally, it rounds the solution via Madow Sampling \cite{Madow1949}.

We target two specific limitations of the MAC framework: its computational cost and its requirement on a user-specified spanning tree ``backbone". Our improvements directly address these challenges to enhance the algorithm's efficiency and generality. Specifically, we focus on 
1) improving computational efficiency and 
2) removing the user-specified backbone requirement.

First, we aim to make MAC faster. The primary computational burden lies in the Frank-Wolfe optimization loop, which is expensive due to the calculation of the Fiedler vector and value (the Fiedler pair) in each iteration. To accelerate MAC, we pursue a two-pronged approach:
\begin{enumerate}
    \item \textbf{Improved Fiedler Pair Calculation} (\Cref{sec:FiedlerValue}): We develop a fast, numerically stable algorithm that leverages the inherent graph structure to calculating the Fiedler vector and value.
    \item \textbf{Improved Convergence Rate} (\Cref{sec:lineSearch}): We investigate step size selection for the Frank-Wolfe algorithm to increase its convergence rate, reducing the total number of iterations required
\end{enumerate}

Second, the original MAC algorithm requires a predefined ``backbone" $\mathcal{E}^f$ to maintain connectivity after rounding (e.g., odometry). We instead consider the case where $\mathcal{E}^f$ is not given or empty, and propose two solutions: first, in \Cref{sec:rounding} we develop a rounding function $\Pi_K$ that returns a connected graph. Alternatively, in \Cref{sec:effR} we propose a heuristic method for automatically determining a strong candidate backbone.

\section{Fiedler Value Calculation} \label{sec:FiedlerValue}
While much of the current research on numerical Fiedler-value solvers \cite{10.5555/1964238.1964283, HSLMC73} focus on scalability to large graphs (exceeding $10^6$ edges), most graphs found in robotics have between $10^3\sim10^5$ edges. At these sizes, we can still utilize the Cholesky Factorization of the \textit{entire} Laplacian under real-time constraints. While it is counterintuitive to tailor our sparsification process to already smaller graphs, it makes our solver well-suited to online, incremental sparsification, where we prune the graph as it grows.
Using this observation, we develop a fast Fiedler value calculation routine, detailed in \Cref{alg:fiedler_shift_invert}.
% \TODO{Write Algorithm for this}

\begin{algorithm}[t!]
\caption{Fiedler Pair Calculation}
\label{alg:fiedler_shift_invert}

\begin{algorithmic}[1]
\Input: $L \in \mathbb{R}^{n \times n} \gets $ Laplacian, $\sigma \in \mathbb{R}\gets$ Shift parameter, $k \in \mathbb{N}\gets$: Number of eigenvalues to compute, $\mathbf{x}_0 \in \mathbb{R}^{n} \gets$: Initial guess vector
\Output: Fiedler Vector $q_2$, Fiedler Value $\lambda_2$
\Function{FiedlerShiftInvert}{$L, \sigma, k, \mathbf{x}_0$}
    \Statex \textbf{Step 1: Shifted System Factorization}
    \State $A \gets L - \sigma I$ \Comment{Shifted matrix}
    \State $p \gets \text{AMD}(A)$ \Comment{Approx. minimum degree \cite{10.1145/1024074.1024081}}
    \State $R \gets \text{Cholesky}(A(p,p))$ \Comment{Cholesky factorization}
    \State $\text{sol}_A(\mathbf{b}) \gets R^{-1}(R^{-\transpose}(\mathbf{b}(p)))$ \Comment{Solves $x$ in $Ax=\mathbf{b}$}

    \Statex \textbf{Step 2: Projected Shift-Invert Operator}
    \State $\mathcal{P}(\mathbf{v}) \gets \mathbf{v} - (\mathbf{1}^\top \mathbf{v}/n) \mathbf{1}$ \Comment{Projection onto $\mathbf{1}^\perp$}
    \State $\mathcal{T}(\mathbf{v}) \gets \mathcal{P}(\text{sol}_A(\mathcal{P}(\mathbf{v})))$ \Comment{Shift-invert operator}

    \Statex \textbf{Step 3: Krylov-Schur Method \cite{doi:10.1137/S0895479800371529}}
    \State $\mathbf{v}_0 \gets \mathcal{P}(\mathbf{x}_0)/\|\mathcal{P}(\mathbf{x}_0)\|$ \Comment{Normalized initial vector}
    \State $m \gets \min(2k + 10, n)$ \Comment{Subspace dimension}
    \State $(\mathbf{V}_m, \mathbf{H}_m) \gets \text{KrylovSchur}(\mathcal{T}, \mathbf{v}_0, m)$
    
    \Statex \textbf{Step 4: Eigenvalue Recovery}
    \State $\lambda_i \gets \sigma + 1/\mathbf{H}_{ii}$ 
    \State Sort $\lambda_i$ ascending, take $\lambda_f \gets \lambda_1$
    \State $\mathbf{x}_f \gets \mathcal{P}(\mathbf{V}_m \mathbf{y}_1)/\|\mathcal{P}(\mathbf{V}_m \mathbf{y}_1)\|$ \Comment{Fiedler vector}

    \State \Return $\mathbf{x}_f, \lambda_f$
\EndFunction
\end{algorithmic}
\end{algorithm}
We leverage three structural properties of the Laplacian. First,  as the Fiedler Value for a Laplacian Graph is the closest eigenvalue to zero, we apply shift-invert conditioning \cite{68344fb6-07ae-3764-b006-3151daf9b1a5}. Next, for the inversion, we perform an approximate minimum degree ordering \cite{10.1145/1024074.1024081} before taking the Cholesky factorization, which, for graphs found in robotics (e.g. pose-graphs), dramatically decreases the number of non-zeros and thus computation time for using our Cholesky factorization. Finally, in our iterative eigenvalue solver we use the fact that the first eigenvalue of our Laplacian will always be zero with an eigenvector of $\mathbf{1}\transpose$, and deflate (project onto $\mathbf{1}^\perp$) our candidate eigenvector $\mathbf{v}$ at each iteration.
To iteratively solve for the eigenvalues of our shift-inverted system, we utilize the Krylov-Schur algorithm \cite{doi:10.1137/S0895479800371529}, and recover our original Fiedler pair with a cheap sort and projection step.

\section{Improving Convergence Rate}\label{sec:lineSearch}
We investigated multiple step size line search and Frank-Wolfe variations for improving the convergence rate of MAC.

Previously, MAC utilized a decaying step size at each iteration:
\begin{equation} \label{eq:decaying}
    \gamma_k  =\frac{2}{2+k}
\end{equation}
Although inexpensive to compute and sharing the same worst-case $O(1/k)$ convergence rate as other step size rules, this simple scheme is often outperformed in practice. In particular, the exact line search and backtracking line search algorithms \cite{pedregosa2020linearly} enjoy linear convergence rates for strongly convex/concave functions with polytope constraints. Although the Fiedler value is not concave, we investigated if the empirical performance gains of both backtracking and exact line search algorithms translate to MAC. The Pair-wise and Away-steps variants of Frank-Wolfe \cite{lacoste2015global} also present similar improvements to the base Frank-Wolfe algorithm. Thus, we implemented MAC for both different line search strategies and Frank-Wolfe variants. Interestingly, as we will discuss in \Cref{sec:LineSearchExp}, Frank-Wolfe with a decaying step size is both the cheapest in terms of total time taken, and the most robust in terms of attaining a higher optimized value. 
\section{Backbone Specification and Removal}\label{sec:Backbone}
One key assumption of MAC is that a fixed set of edges $\mathcal{E}^f$ is given that connects the graph. We improve on this assumption by 1) proposing a method that automatically specifies informative spanning-tree backbones and 2) removing the need for fixed edges entirely. As a result, no user-specified backbone must be given.
\subsection{Rounding with Connectedness} \label{sec:rounding}
A simple backbone-free approach is to solve the relaxed edge selection problem (Problem \ref{prob:relaxedKchoice}) with $\mathcal{E}^f = \emptyset$ and round the solution using the techniques in \cite{doherty2024mac}. However, these rounding methods do not guarantee connectivity, potentially yielding a disconnected graph. To address this, we propose a rounding procedure that ensures connectivity, selects exactly $K$ edges, and maximizes total weight according to the relaxed solution $\bm{x}^*$. Formally, we aim to solve
\begin{equation} \label{eq:RoundFree}
\begin{aligned}
    \max_{\mathbf{w}\in\{0,1\}^m} \quad & \mathbf{w}^\top \bm{x}^* \\
    \text{s.t.} \quad & \sum_{i=1}^m \mathbf{w}_i = K, \quad
    \mathcal{G}(\mathcal{V},\mathcal{E}_{\mathbf{w}})\ \text{connected},
\end{aligned}
\end{equation}
where $\mathcal{E}_\mathbf{w}$ is the set of edges indicated by $\mathbf{w}$. This can be solved efficiently by first computing a maximum spanning tree using $\bm{x}^*$ as weights, then adding the remaining $K-(n-1)$ edges with the highest-weight.

\subsection{Spectrally-Informed Spanning Tree} \label{sec:effR}
Alternatively, rather than using a given set of connected fixed edges, we can automate fixing edges before solving Problem \ref{prob:relaxedKchoice}. However, multiple connected subgraphs or spanning trees may exist for a given graph. Identifying the optimal one requires solving the original unrelaxed problem (Eq. \ref{eq:orioginal_mac}), which is NP-hard; consequently, we rely on heuristic methods to guide the selection.

A natural heuristic for this task is the effective resistance distance, a graph-theoretic quantity that measures the importance of an edge to overall connectivity. Formally, for an edge $(i,j)$, the effective resistance is defined as 
\begin{equation}
    R_{ij}=(\mathbf{e}_i-\mathbf{e}_j)\transpose L^\dagger(\mathbf{e}_i-\mathbf{e}_j)
\end{equation}
where $\mathbf{e}_i\in \mathbb{R}^m$ is the standard basis vector with 1 as its $i^\text{th}$ component, and $L^\dagger$ is the Moore-Penrose pseudo-inverse of $L$ \cite{Klein1993ResistanceD}. Effectively, edges with high effective resistance are bottlenecks—critical connections whose absence would severely disrupt flow. This can arise from an inherently weak connection (low weight) or being a bridge between otherwise disconnected components. We wish to choose a spanning tree that specifically contains edges that are important because of the topology of the graph, and thus we weight each edge's effective resistance by the edge's weight, that is $w_{ij} = R_{ij}w_{ij}$. Then, we determine $\mathcal{E}^f$ as the edges of the maximum spanning tree with weights $w_{ij}$

\sisetup{
  detect-all
}

\definecolor{darkgreen}{rgb}{0.0,0.5,0.0}
\definecolor{darkred}{rgb}{0.6,0,0}

\begin{table}[t!]
\centering
\scriptsize
\begin{tabular}{c l r r | r r r}
% \hline
 & \textbf{Dataset} & \textbf{Nodes} & \textbf{Edges} & \multicolumn{3}{c}{\textbf{Avg. Solve time (ms)}} \\
\cline{5-7}
 &  &  &  & \textbf{Tracemin} & \textbf{Eigs} & \textbf{Ours} \\
\hline
\multirow{3}{*}{\rotatebox[origin=c]{90}{\textbf{SFM}}} 
 & IMC-gate       &  5436  & 216985 & \makebox[4em][r]{263 (\textcolor{darkgreen}{-85\%})} & 1739 & \makebox[4em][r]{\textbf{65 (\textcolor{darkgreen}{-96\%})}} \\
 & BAL-392       & 28713  & 128565 & \makebox[4em][r]{177 (\textcolor{darkgreen}{-8.0\%})} & 192 & \makebox[4em][r]{\textbf{21 (\textcolor{darkgreen}{-89\%})}} \\
 & Replica-office &  1989  &  63921 & \makebox[4em][r]{290 (\textcolor{darkred}{675\%})} & 37 & \makebox[4em][r]{\textbf{10 (\textcolor{darkgreen}{-73\%})}} \\
\hline
\multirow{3}{*}{\rotatebox[origin=c]{90}{\textbf{SNL}}} 
 & sphere2500-snl &  2500  &  4949  & \makebox[4em][r]{11 (\textcolor{darkred}{270\%})} & \textbf{3} & \makebox[4em][r]{4 (\textcolor{darkred}{42\%})} \\
 & city10k-snl    & 10000  & 20687  & \makebox[4em][r]{41 (\textcolor{darkred}{109\%})} & 20 & \makebox[4em][r]{\textbf{14 (\textcolor{darkgreen}{-31\%})}} \\
 & grid3D-snl     &  8000  & 22236  & \makebox[4em][r]{88 (\textcolor{darkred}{68\%})} & 53 & \makebox[4em][r]{\textbf{39 (\textcolor{darkgreen}{-25\%})}} \\
\hline
\multirow{4}{*}{\rotatebox[origin=c]{90}{\textbf{PGO}}} 
 & sphere2500     &  2500  &  4949  & \makebox[4em][r]{12 (\textcolor{darkred}{127\%})} & 5 & \makebox[4em][r]{\textbf{5 (\textcolor{darkgreen}{-14\%})}} \\
 & city10000      & 10000  & 20687  & \makebox[4em][r]{41 (\textcolor{darkred}{199\%})} & 14 & \makebox[4em][r]{\textbf{14 (\textcolor{darkgreen}{-0.7\%})}} \\
 & grid3D         &  8000  & 22236  & \makebox[4em][r]{74 (\textcolor{darkgreen}{-12\%})} & 85 & \makebox[4em][r]{\textbf{39 (\textcolor{darkgreen}{-54\%})}} \\
  & tiers         &  9769  & 17553  & \makebox[4em][r]{123 (\textcolor{darkred}{288\%})} & 31 & \makebox[4em][r]{\textbf{15 (\textcolor{darkgreen}{-52\%})}} \\
\hline
\end{tabular}
\caption{\textbf{Eigenvalue Solve Time}. Frank–Wolfe is run on sparsified SfM, PGO, and SNL graphs using \textit{Tracemin}, \textit{Eigs}, and \textit{Our} solver; times are reported with percent change ($\frac{t-t_{\textit{Eigs}}}{t_{\textit{Eigs}}}$) against \textit{Eigs} (\textcolor{darkgreen}{faster}, \textcolor{darkred}{slower}), \textbf{bold} marks the fastest.}
\label{tab:solve_times}
\squeezeup
\end{table}
\begin{figure}[t!]
\centering
\vspace{0.5em} % small vertical space between rows
% % Row 2
\begin{subfigure}[b]{0.30\textwidth}
    \centering
    \includegraphics[width=\linewidth]{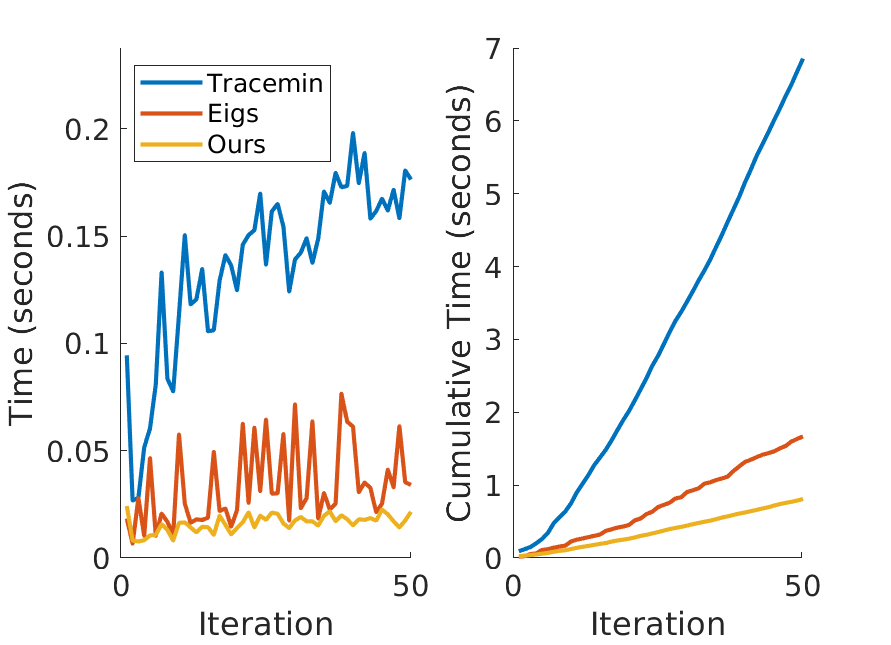}
    % \caption{}
\end{subfigure}
\hfill
\begin{subfigure}[b]{0.30\textwidth}
    \centering
    \includegraphics[width=\linewidth]{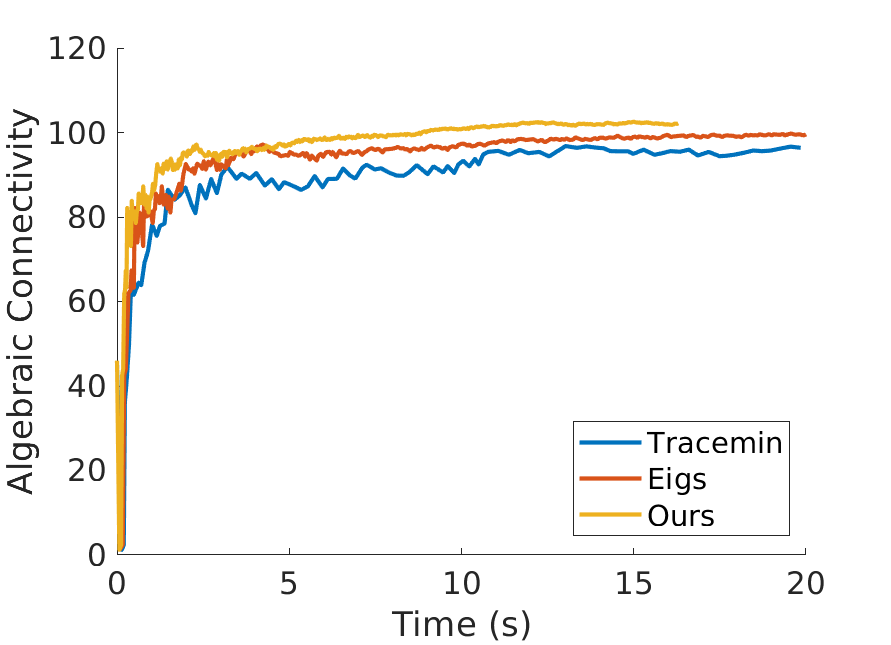}
    % \caption{}
\end{subfigure}
\caption{\textbf{TIERS} \cite{yu2023fusing} dataset sparsified to 80\% edges via Frank-Wolfe. Notably, each iteration takes significantly less time when using our algorithm due to the faster Fiedler Pair solves (\textbf{top}), resulting in higher algebraic connectivity $\lambda_2$ in shorter time (\textbf{bottom})}
\label{fig:EigSolvetimes}
\squeezeup
\end{figure}
\section{Experimental Results}
We evaluated our algorithms on variety of datasets representing three common estimation problems: Pose Graph Optimization (PGO), Sensor Network Localization (SNL), and Structure from Motion (SfM). Experiments were conducted on both synthetic and real datasets (results on a wider range of datasets can be found in the supplementary material). To generate sensor network localization graphs, pose graphs were converted by replacing poses with landmarks, and transformations as range measurements. We seek to answer several questions:
\begin{itemize}
    \item \textbf{Q1}: Can MAC's runtime be improved through specialized eigenvalue solvers?
    \item \textbf{Q2}: How do different types of graphs found in robotics affect MAC?
    \item \textbf{Q3}: Do the advanced variants of Frank-Wolfe result in fewer iterations to solve Problem \ref{prob:relaxedKchoice}?
    \item \textbf{Q4}: How can we automatically choose a high quality ``backbone"?
    \item \textbf{Q5}: How does the Fiedler value of a graph produced by our rounding scheme compare to that of a graph with a pre-specified backbone?
\end{itemize}

\subsection{Fiedler Value Calculation} \label{sec:FiedlerExp}
To evaluate \Cref{alg:fiedler_shift_invert}, we compared our algorithm (\textit{Ours}) against two established solvers: the Tracemin-Fiedler algorithm (\textit{Tracemin}) with Cholesky preconditioning \cite{10.5555/1964238.1964283}, and \texttt{eigs}, the Krylov–Schur method (\textit{Eigs}) available in MATLAB. Our setup began by running MAC \cite{doherty2024mac} for 50 iterations on each dataset (presented: IMC-gate \cite{image-matching-challenge-2023}, BAL-392 \cite{agarwal2010bundle}, Replica \cite{straub2019replica}, Sphere, Grid \cite{7139836}, and City\cite{kaess2008isam}) and saving the Laplacians produced at every iterate. We then applied each solver to these Laplacians and recorded runtimes. As reported in \Cref{tab:solve_times}, our method consistently achieved the fastest performance across datasets, with the sole exception of the Sphere2500-SNL dataset.  

To test our hypothesis that eigenvalue computation is the dominant cost in Frank–Wolfe, we next integrated each solver into a Frank–Wolfe routine. The results, summarized in \Cref{fig:EigSolvetimes}, show that our algorithm again outperformed both Tracemin and Eigs, confirming that our approach substantially reduces the overall runtime (\textbf{Q1}). Additionally, our algorithm demonstrates consistently better performance across all three different types of graphs (\textbf{Q2}).

\subsection{Line Search} \label{sec:LineSearchExp}
We next examined the effect of line-search strategies and Frank–Wolfe variants on several real datasets, including TIERS \cite{yu2023fusing}, MR.CLAM7 \cite{10.1177/0278364911398404}, and Intel \cite{Carlone2014AFA}. Alongside the simple (\textit{Naive}) decaying step size \eqref{eq:decaying} which requires no additional objective evaluations, we implemented an approximate exact line search (\textit{Exact}) using the \texttt{fminbnd} function in MATLAB. 
Since this approach can be computationally expensive, we also considered an approximate backtracking line search (\textit{Backtracking}) following \cite{pedregosa2020linearly}. Finally, we implemented the pairwise Frank–Wolfe algorithm (\textit{PFW-Exact})\cite{lacoste2015global}. Of note, across our datasets shown here, the away-step is never taken for an away-steps Frank-Wolfe and thus we do not include it in our results.
\begin{figure*}[tbp]
\centering

% % Row 2
\begin{subfigure}[b]{0.3\textwidth}
    \centering
    \includegraphics[width=\linewidth]{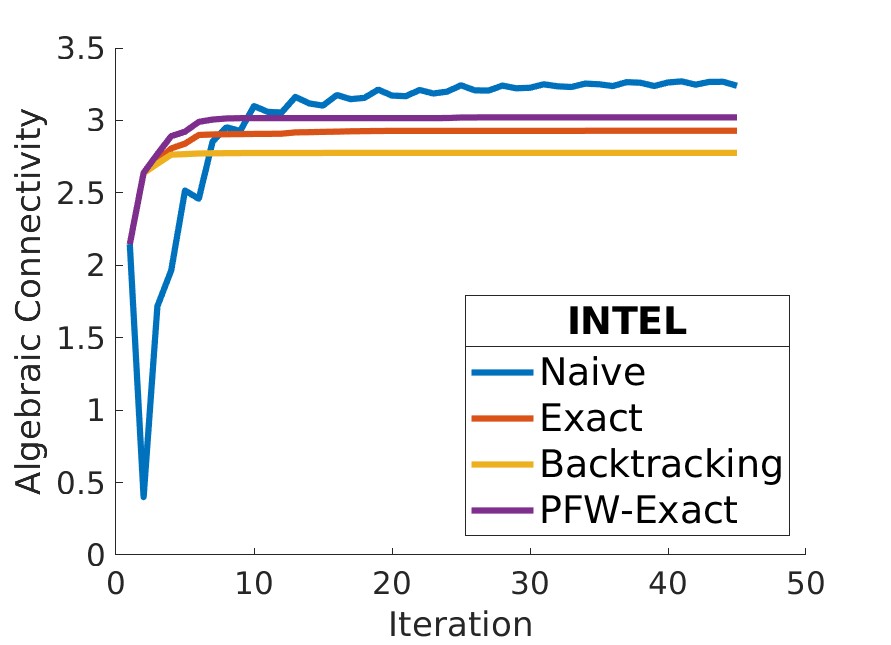}
    % \caption{Intel}
\end{subfigure}
\hfill
\begin{subfigure}[b]{0.3\textwidth}
    \centering
    \includegraphics[width=\linewidth]{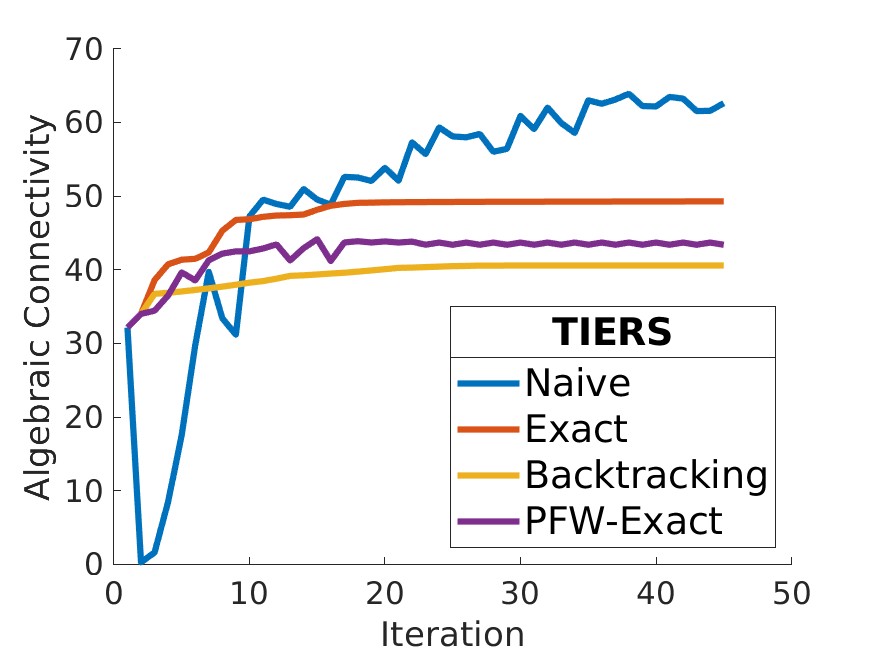}
    % \caption{MRCLAM7}
\end{subfigure}
\hfill
\begin{subfigure}[b]{0.3\textwidth}
    \centering
    \includegraphics[width=\linewidth]{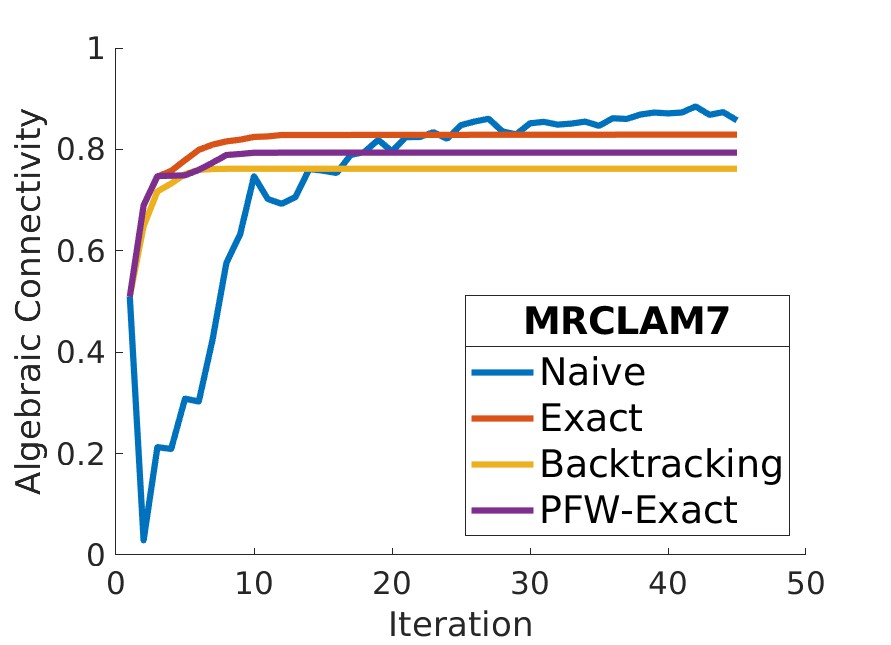}
    % \caption{TIERS}
\end{subfigure}

\caption{\textbf{Algebraic Connectivity vs Iteration}: Each graph corresponds to the runs in \Cref{tab:linesearchTimes}: As seen, the \textit{Exact, Backtracking, and PFW-Exact} variants stall in optimization progress in a few iterates. While the \textit{Naive} step size strategy initially takes a poor step, it does not suffer from this stalling behavior.}
\label{fig:linesearch}
\end{figure*}

The results, shown in \Cref{tab:linesearchTimes} and \Cref{fig:linesearch}, reveal that the decaying step size not only achieved the lowest runtime but also produced the highest Fiedler value. In contrast, the other methods took good initial iterates, but stalled later at suboptimal values. 
% Although the decaying step size initially took poor steps, it ultimately avoided premature stagnation and surpassed the alternatives in terms of $\lambda_2$.  
Thus, the naive Frank-Wolfe proves to be both faster and produce better iterates than the more advanced variants (\textbf{Q3}). 

We believe this outcome stems from the nonsmooth nature of the Fiedler value function. At edge weights where the Laplacian exhibits repeated Fiedler eigenvalues, the gradient is undefined. While our supergradient still produces a valid descent direction, it may provide little useful information in these cases. As mentioned, for most datasets very few away-steps are taken when using the away Frank-Wolfe, indicating our first couple iterates move towards a ``good" set of faces of our constraint set. However, moving from this ``good'' set of faces to the true optimal set of faces becomes difficult with a conservative line search algorithm along an arbitrary supergradient. In contrast, the decaying step size produces a zig-zagging effect which, while usually undesirable, allows the exploration of more faces.

\sisetup{round-mode=places,round-precision=3}

\begin{table}[t!]
\centering
\scriptsize
\renewcommand*{\arraystretch}{1.1}
\begin{tabular}{@{}l l *{4}{c}@{}}
% \toprule
 & & \multicolumn{4}{c}{Line Search Type} \\
\cmidrule(lr){3-6}
 & Metric & Naive & Exact & Backtracking & PFW-Exact \\
\midrule
\multirow{3}{*}{\adjustbox{valign=m}{\rotatebox{90}{\textbf{Intel}}}}
% \multirow{3}{*}{\rotatebox[origin=c]{90}{\makebox[2.2cm][c]{input-INTEL}}}
& Cum. time (s)   & \textbf{0.138} & 1.064 & 0.140 & 1.006 \\
& $\lambda_2$      & \textbf{3.260} & 2.929 & 2.753 & 3.021\\
& Avg time (s)    &  0.003 & 0.024 & \textbf{0.003} & 0.022\\
\midrule
\multirow{3}{*}{\adjustbox{valign=m}{\rotatebox{90}{\textbf{TIERS}}}} 
& Cum. time (s)   & \textbf{1.153} & 19.317 & 1.889 & 14.190 \\
& $\lambda_2$      & \textbf{63.573} & 49.294 & 42.718 & 43.699 \\
& Avg time (s)    & \textbf{0.026} & 0.429 & 0.042 & 0.315 \\
\midrule
\multirow{3}{*}{\adjustbox{valign=m}{\rotatebox{90}{\textbf{MRCLAM7}}}} 
& Cum. time (s)   & \textbf{2.342 }& 32.551 & 3.210 & 35.235 \rule{0pt}{2.9ex}\\
& $\lambda_2$      & \textbf{0.873} & 0.829 & 0.743 & 0.794 \rule{0pt}{2.9ex}\\
& Avg time (s)    & \textbf{0.052} & 0.723 & 0.065 & 0.782 \rule{0pt}{2.9ex}\\
\addlinespace[1.2ex]
\bottomrule
\setlength{\extrarowheight}{0pt}
\end{tabular}
\caption{\textbf{Frank-Wolfe Variants}. We report cumulative runtime, the resulting Fiedler value ($\lambda_2$), and average runtime per iteration on a single run of each dataset. \textbf{Bold} indicates the best solver. Surprisingly, the simple decaying stepsize (\textit{Naive}) consistently outperforms the more advanced variants
}
\label{tab:linesearchTimes}
\squeezeup
\end{table}

\subsection{Backbone} \label{sec:BackboneExp}

\begin{figure*}[tbp]
\centering

\vspace{0.5em} % small vertical space between rows

% % Row 2
\begin{subfigure}[b]{0.3\textwidth}
    \centering
    \includegraphics[width=\linewidth]{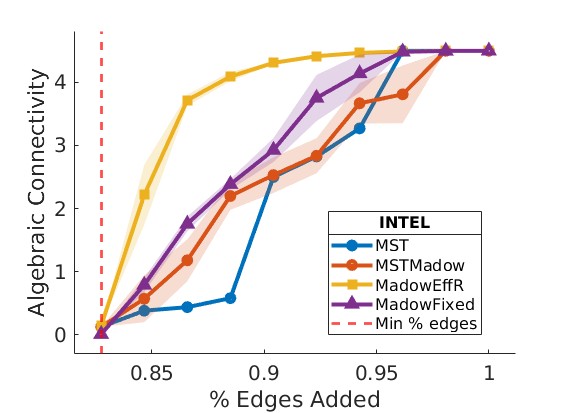}
    % \caption{}
\end{subfigure}
\hfill
\begin{subfigure}[b]{0.3\textwidth}
    \centering
    \includegraphics[width=\linewidth]{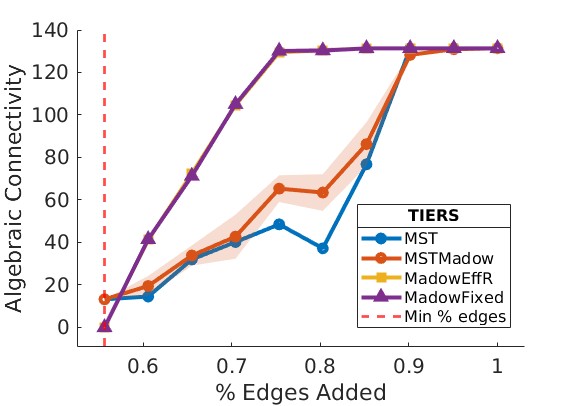}
    % \caption{}
\end{subfigure}
\hfill
\begin{subfigure}[b]{0.3\textwidth}
    \centering
    \includegraphics[width=\linewidth]{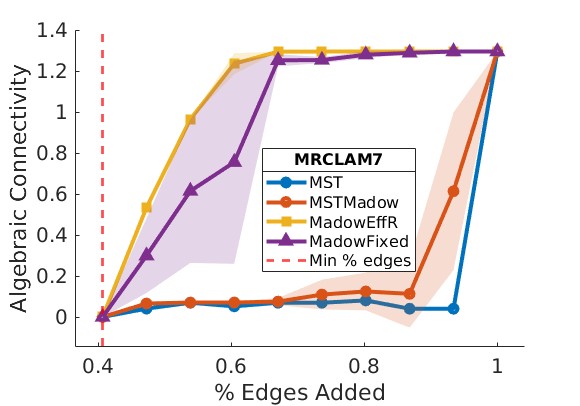}
    % \caption{}
\end{subfigure}

\vspace{0.5em}

% Row 3
\begin{subfigure}[b]{0.3\textwidth}
    \centering
    \includegraphics[width=\linewidth]{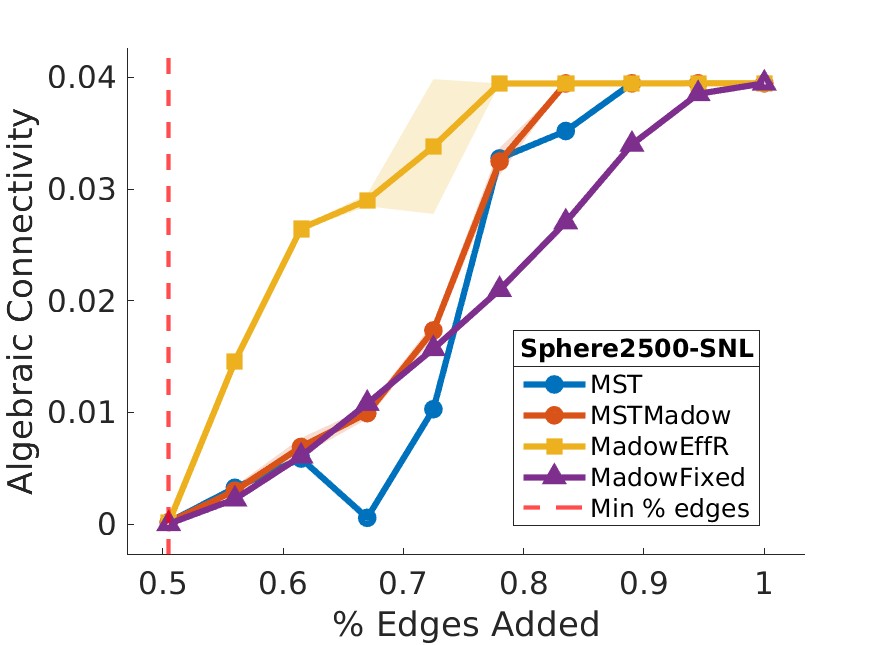}
   
    % \caption{}
\end{subfigure}
\hfill
\begin{subfigure}[b]{0.3\textwidth}
    \centering
    \includegraphics[width=\linewidth]{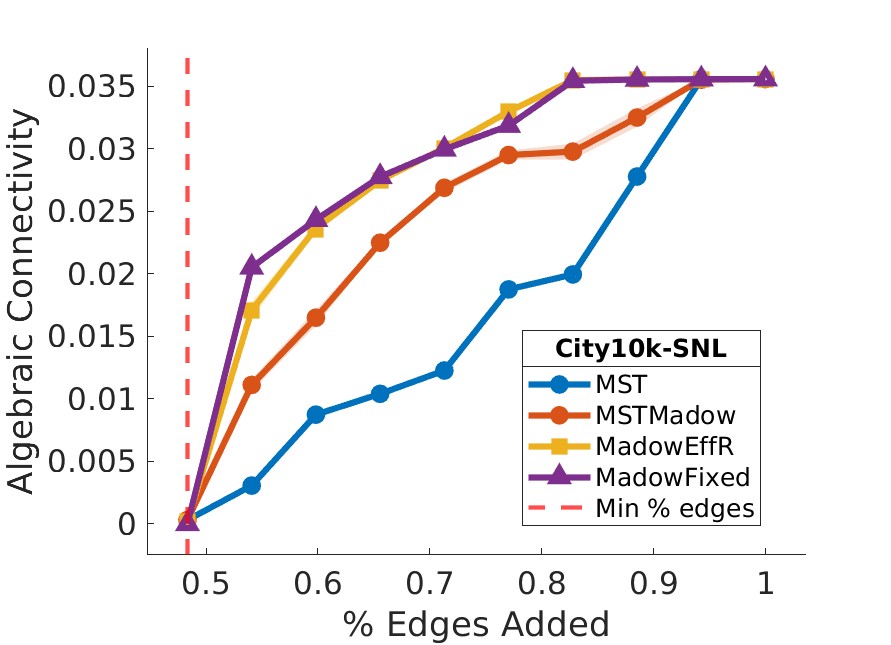} 
    % \caption{}
\end{subfigure}
\hfill
\begin{subfigure}[b]{0.3\textwidth}
    \centering
    \includegraphics[width=\linewidth]{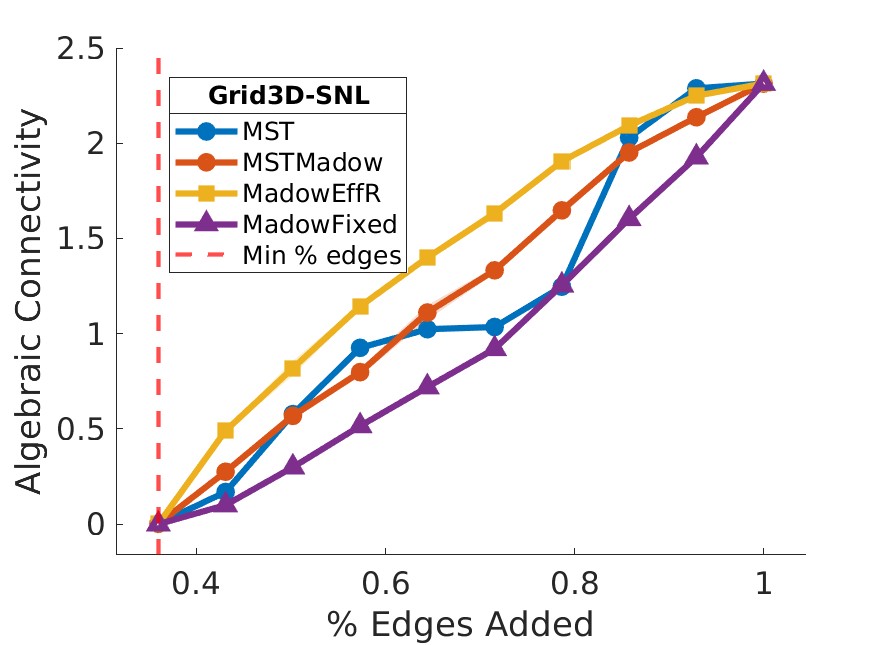}
    % \caption{}
\end{subfigure}

\caption{\textbf{Algebraic Connectivity vs Edge Budget}. Graphs of $\lambda_2$ vs \% edges kept. Shaded error bars indicate the one Std Dev. bound around the mean cost for Madow rounding strategies. Our heuristic backbone generally outperforms all other strategies, while interestingly the backbone-free approaches (MST, MSTMadow) generally perform the worst.}
\label{fig:Backbone}
\squeezeup
\end{figure*}

Our final set of experiments compared rounding and spanning-tree strategies. We considered several approaches varying backbone and rounding schemes. First, the \textit{MST} strategy, described in \Cref{sec:rounding}, solves using an empty fixed edge set. A second variant, \textit{MST-Madow}, augments \textit{MST} by using Madow sampling \cite{Madow1949} to choose the $K - (n-1)$ edges after finding a max spanning tree. We also evaluated \textit{MadowEffR}, which combines the spectral effective resistance backbone from \Cref{sec:Backbone} with Madow sampling. For comparison with prior work, we reproduced the odometry-based fixed backbone of \cite{doherty2024mac} (\textit{MadowFixed}), again coupled with Madow rounding. Because Madow rounding is randomized, each experiment was repeated ten times, and we report both the mean and one-standard-deviation bounds. A subtlety of these comparisons is that MST-based methods must solve a higher-dimensional Frank–Wolfe problem, since all edges are initially treated as unfixed.  

As shown in \Cref{fig:Backbone}, the effective-resistance backbone generally outperforms the alternatives, with only the odometry backbone producing similar performance on select datasets. Thus, our backbone specification heuristic empirically returns stronger sets of fixed edges (\textbf{Q4}). Surprisingly, the MST-based methods perform poorly across datasets, despite explicitly selecting spanning-tree weights that maximize the Fiedler value (\textbf{Q5}). We hypothesize this is because the relaxed problem has many symmetries (see \cite[Sec.~V.A]{doherty2024mac}), and thus solving the relaxed problem (Problem \ref{prob:relaxedKchoice}) spreads the edge selection weights across many potential subgraphs. When rounding on these edge selection weights, we return a mediocre graph. In contrast, by specifying a connected subgraph \textit{before} optimization, we break this symmetry.

\section{Conclusion}
In this paper, we propose several improvements to the Maximizing Algebraic Connectivity (MAC) algorithm. In particular, we significantly enhance computational efficiency by accelerating the Fiedler value computation through a specialized solver. We also uncover a surprising result: the naive step size strategy is not only the fastest and most computationally inexpensive approach, but it also achieves higher Fiedler values than more sophisticated alternatives. Finally, we present methods to guarantee connectedness in the returned graph and introduce a spectral-based heuristic that consistently outperforms the original MAC algorithm.

One exciting direction for future work lies in understanding why the naive step size strategy outperforms more advanced Frank–Wolfe variants. A deeper analysis of the Fiedler value optimization landscape may yield insights that inspire more effective first-order solvers. Another promising avenue is to better understand why specifying a backbone \emph{a priori} leads to markedly improved results and whether we can formally characterize the error introduced by such a specification. For example, one might solve MAC with our rounding scheme, fix a subset of edges, then re-solve MAC on the remaining graph iteratively. Determining whether this approach yields provably stronger solutions would be an important step forward.

Overall, our findings not only improve the state of the art in maximizing algebraic connectivity but also highlight several intriguing theoretical questions. By bridging the gap between practical heuristics and rigorous analysis, future work has the potential to further advance both the efficiency and reliability of connectivity optimization algorithms.
% \section*{Acknowledgment}

\begingroup
\bibliographystyle{IEEEtran}
\scriptsize
\bibliography{refs}
\endgroup

\end{document}